\documentclass[10pt,twocolumn,letterpaper]{article}

\usepackage{cvpr}
\usepackage{times}
\usepackage{epsfig}
\usepackage{graphicx}
\usepackage{amsmath}
\usepackage{amssymb}
\usepackage{footnote}
\usepackage{dsfont}

\usepackage{comment}
\usepackage{blindtext}
\usepackage{morewrites}
\usepackage{makecell}
\usepackage{pifont}
\newcommand{\xmark}{\ding{55}}%
\newcommand{\cmark}{\ding{51}}%
\usepackage{multirow}

\makeatletter
\renewcommand{\paragraph}{%
	\@startsection{paragraph}{4}%
	{\z@}{1ex \@plus 1ex \@minus .2ex}{-1em}%
	{\normalfont\normalsize\bfseries}%
}
\makeatother

\usepackage[breaklinks=true,bookmarks=false]{hyperref}

\cvprfinalcopy 

\def\cvprPaperID{****} 

\ifcvprfinal\fi
\begin{document}

\title{Vec2Face: Unveil Human Faces from their Blackbox Features in \\ Face Recognition}

\author{Chi Nhan Duong$^{1}$, Thanh-Dat Truong$^{2}$, Kha Gia Quach$^{1}$, Hung Bui$^{3}$, Kaushik Roy$^{4}$, Khoa Luu$^{2}$\\
    $^{1}$ Concordia University, Canada \quad 
	$^{2}$ University of Arkansas, USA \quad
	$^{3}$ VinAI Research \\ 
	$^{4}$ North Carolina A\&T State University, USA \\
	\tt\small $^{1}$\{dcnhan, kquach\}@ieee.org, $^{2}$\{tt032, khoaluu\}@uark.edu,  \tt\small$^{3}$v.hungbh1@vinai.io,   $^{4}$kroy@ncat.edu
}

\maketitle


\begin{abstract}
   Unveiling face images of a subject given his/her high-level representations extracted from a blackbox Face Recognition engine is extremely challenging. It is because the limitations of accessible information from that engine including its structure and uninterpretable extracted features. This paper presents a novel generative structure with Bijective Metric Learning, namely Bijective Generative Adversarial Networks in a Distillation framework (DiBiGAN), for synthesizing  faces of an identity given that person's features. In order to effectively address this problem, this work firstly introduces a bijective metric so that the distance measurement and metric learning process can be directly adopted in image domain for an image reconstruction task. Secondly, a distillation process is introduced to maximize the information exploited from the blackbox face recognition engine. Then a Feature-Conditional Generator Structure with Exponential Weighting Strategy is presented for a more robust generator that can synthesize realistic faces with ID preservation. Results on several benchmarking datasets including CelebA, LFW, AgeDB, CFP-FP against matching engines have demonstrated the effectiveness of DiBiGAN on both image realism and ID preservation properties. 
\end{abstract}

\vspace{-5mm}
\section{Introduction}

Face recognition has recently matured and achieved high accuracy against millions of identities \cite{Xu_TIP2015, Xu_IJCB2011}. A face recognition system is often designed in two main stages, i.e. feature extraction and feature comparison. 
The role of feature extraction is more important since it directly determines the robustness of the engine. 
This operator defines an embedding process mapping input facial images into a higher-level latent space where embedded features extracted from photos of the same subject distribute within a small margin \cite{Luu_IJCB2011}.
Moreover, since most face recognition engines are set into a blackbox mode to protect the technologies \cite{Le_JPR2015}, there is no apparent technique to inverse that embedding process to reconstruct the faces of a subject given his/her extracted features from those engines.

\begin{figure}[t]
	\centering \includegraphics[width=0.95\columnwidth]{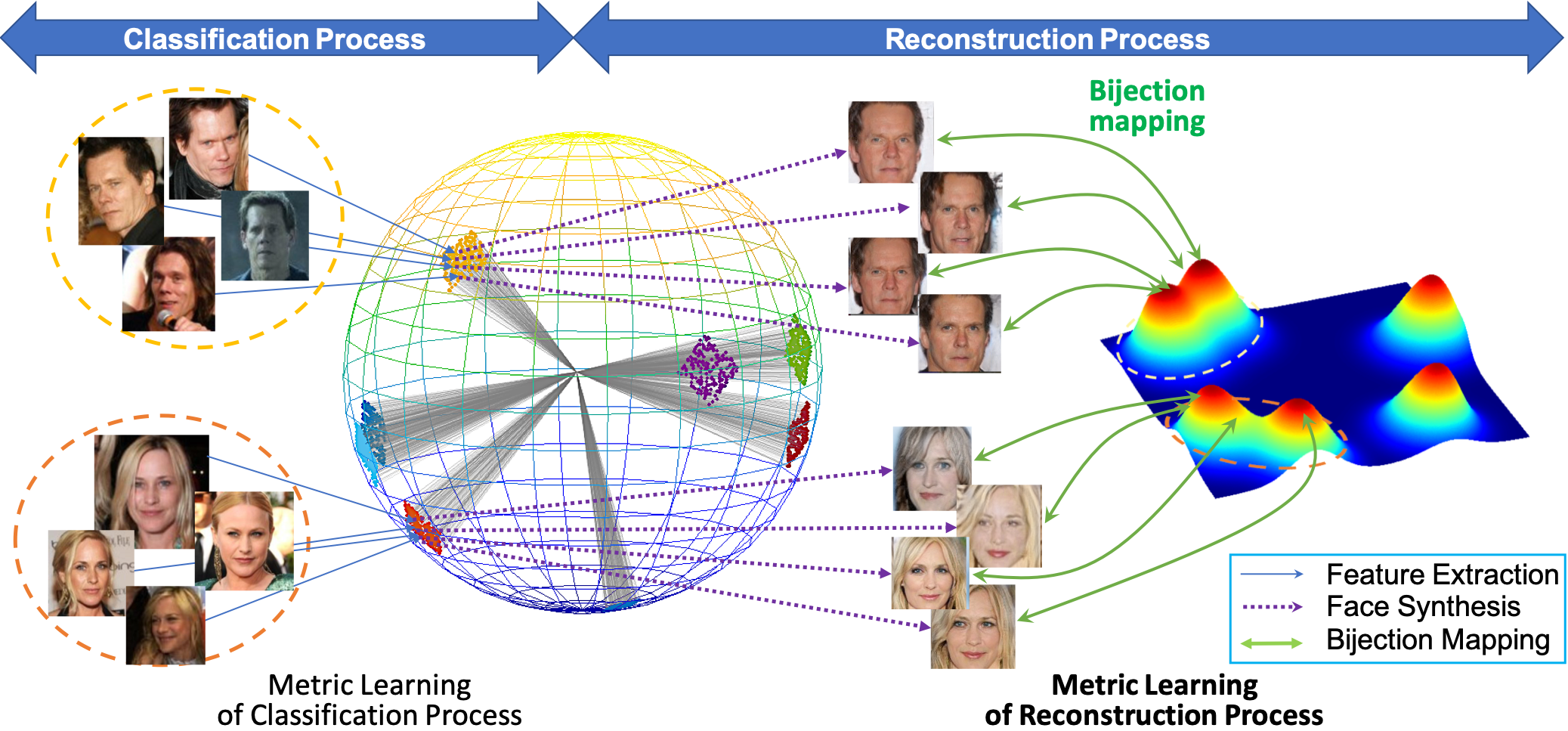}
	\small
	\caption{\textbf{Metric Learning for Image Reconstruction.} By maintaining the one-to-one mapping via a bijection, the distance between images can be directly and intuitively measured and enhances the metric learning process for image reconstruction task.}
	\label{fig:BijectiveMetricLearningIllutration}
	\vspace{-5mm}
\end{figure}

\begin{table*} [!t]
	\small
	\centering
	\caption{Comparisons of our DibiGAN and other unrestricted synthesis methods. Image Reconstruction (Img\_Recon), Feature Representation (Feat), Guided Image (Img$_G$), Feature Conditional (Feat\_Cond), Neighborly Deconvolution (NB\_Deconv), Optimization (Opt).} 
	\footnotesize
     \begin{tabular}{ >{\arraybackslash}m{2.9cm} c c c c c} 
		\Xhline{2\arrayrulewidth}
		& \textbf{Ours} & NBNet \cite{mai2018reconstruction}  & SynNormFace \cite{Cole_2017_CVPR}  & IFaceRec  \cite{Zhmoginov2016InvertingFE} & INVREP \cite{conf/cvpr/MahendranV15} \\
		
		\Xhline{2\arrayrulewidth}
		
		\textbf{Input} & Feat & Feat  & Feat & Feat + Img$_G$ & Feat \\ 
		\begin{tabular}{@{}l@{}}\textbf{Generator}  \textbf{Structure} \end{tabular}   & \begin{tabular}{@{}c@{}}\textbf{Feat\_Cond} \end{tabular} & \begin{tabular}{@{}c@{}}NB\_Deconv \end{tabular}  & \begin{tabular}{@{}c@{}} MLP + CNN \end{tabular}& DeConvNet & Opt \\
		
		\textbf{Blackbox Support} & \cmark & \cmark  & \xmark & \xmark & \xmark \\
		\hline
		\begin{tabular}{@{}l@{}}\textbf{Img\_Recon Metric} \end{tabular}& \begin{tabular}{@{}l@{}} \textbf{Bijective} \end{tabular}& \xmark & \xmark & \xmark & \xmark \\ 
		\textbf{Exploited Knowledge} \textbf{from Classifier}& \begin{tabular}{@{}c@{}}\textbf{Fully}\\\textbf{(Distillation)} \end{tabular}& \begin{tabular}{@{}l@{}}Partially \end{tabular}& \begin{tabular}{@{}c@{}}Fully\\(Whitebox) \end{tabular}& \begin{tabular}{@{}c@{}}Fully\\(Whitebox) \end{tabular} & \begin{tabular}{@{}c@{}}Fully\\(Whitebox) \end{tabular} \\ 
		
		\Xhline{2\arrayrulewidth}
		
	\end{tabular}\label{tab:TenMethodSumm}
	\vspace{-4mm}
\end{table*}

Some Blackbox Adversarial Attack approaches \cite{pmlr-v80-ilyas18a,ilyas2018prior,Thys_2019_CVPR_Workshops} have partially addressed this task by analyzing the gradients of the classifier's outputs to generate adversarial examples that mislead the behaviour of that classifier. However, they only focus on a \textit{closed-set} problem where the output classes are predefined. Moreover, their goal is to generate imperceptible pertubations added to the given input signal. Other methods \cite{pmlr-v80-athalye18b, Cole_2017_CVPR,Dosovitskiy_2016_CVPR, NIPS2018_8052,Zhmoginov2016InvertingFE} are also introduced in literature but still require the access to the classifier structure, i.e. whitebox setting. Meanwhile, our goal focuses on a more challenging reconstruction task with a \textit{blackbox} face recognition. Firstly, this process \textit{reconstructs faces from scratch} without any hint from input images. 
Secondly, in a blackbox setting, there is \textit{no information about the engine's structure}, and, therefore, it is unable to directly exploit knowledge from the inverse mapping process (i.e. back-propagation). 
Thirdly, the embedded features from a face recognition engine are for \textit{open-set problem} where no label information is available. More importantly, the subjects to be reconstructed may have never been seen during training process of the face recognition engine. In the \textit{scope of this work}, we assume that the face recognition engines are primarily developed by Convolutional Neural Networks (CNN) that dominate recent state-of-the-art results in face recognition \cite{deng2019arcface,duong2019shrinkteanet, duong2018mobiface, LiuNIPS18,liu2017sphereface, schroff2015facenet, wang2018cosface,wen2016discriminative, zhang2019adacos}. We also assume that there is no further post-processing after the step of CNN feature extraction. 
We then develop a theory to guarantee the reconstruction robustness of the proposed method.

\noindent
\textbf{Contributions.} This paper presents a novel generative structure, namely Bijective Generative Adversarial Networks in a Distillation framework (DiBiGAN), with Bijective Metric Learning for the image reconstruction task. The contributions of this work are four-fold. 
(1) Although many metric learning techniques have been introduced in the literature, they are mainly adopted for classification rather than \textit{reconstruction} process. 
By addressing limitations of classifier-based metrics for image reconstruction, we propose a novel \textit{\textbf{Bijective Metric Learning}} with bijection (one-to-one mapping) property so that the distances in latent features are equivalent to those between images (see Fig. \ref{fig:BijectiveMetricLearningIllutration}). It, therefore,  provides a more effective and natural metric learning approach to the image reconstruction task.
(2) We exploit different aspects of the \textit{distillation process} for the image reconstruction task in a blackbox mode. They include \textit{distilled knowledge} from the blackbox face matcher and \textit{ID knowledge extracted} from a real face structure. 
(3) We introduce a Feature-Conditional Generator Structure with Exponential Weighting Strategy for Generative Adversarial Network (GAN)-based framework to learn a more robust generator to synthesize realistic faces with ID preservation.
(4) Evaluations on benchmarks against various face recognition engines have illustrated the improvements of DiBiGAN in both image realism and ID preservation. 
To the best of our knowledge, this is one of the first metric learning methods for image reconstruction (Table \ref{tab:TenMethodSumm}).

\begin{figure*}[t]
	\centering \includegraphics[width=1.95\columnwidth]{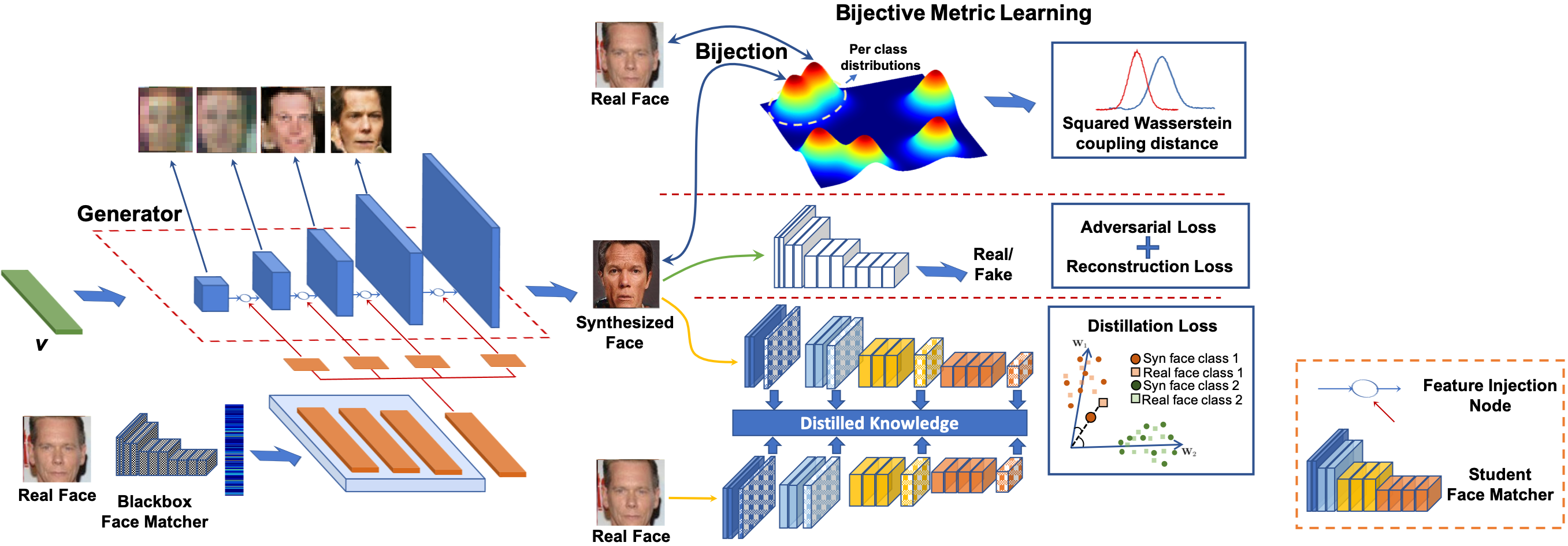}
	\caption{\textbf{Proposed Framework.} Given a high-level embedding representation,
		a \textit{Feature-Conditional Generator} injects that representation through-out its structure as the conditional information for all scales. The cost functions are designed with \textit{Bijective Metric} to directly exploit ID distributions in image domain, and \textit{Distillation Loss} to maximize the knowledge could be extracted from the blackbox matcher.}
	\label{fig:ProposedFramework}
	\vspace{-4mm}
\end{figure*}

\section{Related Work}
\vspace{-1mm}
Synthesizing images \cite{Cole_2017_CVPR,Dosovitskiy_2016_CVPR,nhan2015beyond,duong2019deep,mai2018reconstruction,Zhmoginov2016InvertingFE}  has brought several interests from the community. 
We divided into two groups, i.e. unrestricted and adversarial synthesis. 

\noindent
\textbf{Unrestricted synthesis.} The approaches focus on reconstructing an image from scratch given its high-level representation. 
Since the mapping is from a low-dimensional latent space to a highly nonlinear image space, several regularizations have to be applied, e.g. Gaussian Blur \cite{yosinski2015understanding} for high-frequency samples or Total Variation \cite{duong2019automatic, conf/cvpr/MahendranV15} for maintaining piece-wise constant patches. These optimization-based techniques are  limited with high computation or unrealistic reconstructions. Later, Dosovitskiy et al. \cite{Dosovitskiy_2016_CVPR} proposed to reconstruct the image from its shallow (i.e. HOG, SIFT) and deep features using a Neural Network (NN). Zhmoginov et al. \cite{Zhmoginov2016InvertingFE} presented an iterative method to invert Facenet \cite{schroff2015facenet} feature with feed-forward NN. Cole et al. \cite{Cole_2017_CVPR} proposed an autoencoder structure to map the features to frontal neutral face of the subject. Yang et al. \cite{Yang:2019:NNI:3319535.3354261} adopted autoencoder for model inversion task.
Generally, to produce better synthesized quality, these approaches require full access to the deep structure to exploit the gradient information from the embedding process.
Mai et al. \cite{mai2018reconstruction} developed a neighborly deconvolutional network to support the blackbox mode. However, with only pixel and perceptual \cite{johnson2016perceptual} losses, there are limitations of ID preservation when synthesizing different features of the same subject. In this work, we address this issue with Bijective Metric Learning and Distillation Knowledge for reconstruction task. 

\noindent
\textbf{Adversarial synthesis.} 
Adversarial approaches aim at generating unnoticable perturbations from input images for adversarial examples to mislead the behaviour of a deep structure. 
Either directly accessing or indirectly approximating gradients, adversarial examples are created by maximizing corresponding loss which can fool a classifier \cite{pmlr-v80-athalye18b, Brunner2019CopyAP, Cheng2019ImprovingBA, pmlr-v80-ilyas18a, ilyas2018prior,   Liu2016DelvingIT,MoosaviDezfooli2015DeepFoolAS, Shukla2019BlackboxAA, Thys_2019_CVPR_Workshops}. Ilyas et. al. \cite{ilyas2018prior} proposed bandit optimization to exploit prior information about the gradient of deep learning models.  Later, Ilyas et. al. \cite{pmlr-v80-ilyas18a} introduced Natural Evolutionary Strategies to enable query-efficient generation of black-box adversarial examples.
Other knowledge from the blackbox classifier are also exploited for this task \cite{Thys_2019_CVPR_Workshops, pmlr-v80-athalye18b, NIPS2018_8052}.
Generally, although the approaches in this direction tried to extract the gradient information from a blackbox classifier, their goal are mainly to mislead the behaviours of the classifier with respect to a pre-defined set of classes. Therefore, they are closed-set approaches. Meanwhile, in our work, the proposed framework can reconstruct faces of subjects that have not been seen in the training process of the classifier.

\vspace{-2mm}
\section{Our Proposed Method}

Let $F: \mathcal{I} \mapsto \mathcal{F}$ be a function that maps an input image $I$ from image domain $\mathcal{I} \in \mathbb{R}^{W \times H \times C}$ to its high-level embedding feature $F(I)$ in latent domain $\mathcal{F} \in \mathbb{R}^M$. In addition, a function $C: \mathcal{F} \mapsto \mathcal{Y}$ 
takes $\mathcal{F}(I)$ as its input and gives the identity (ID) prediction of the subject in space $\mathcal{Y} \in \mathbb{R}^N$ where each dimension represents a predefined subject class.

\noindent
\textbf{Definition 1} (Model Inversion). \textit{Given blackbox functions $F$ and $C$; and a prediction score vector $s = [F \circ C] (I)$ extracted from an unknown image $I$, the goal of model inversion is to recover $I$ from $s$ such that $\tilde{I}^* = \arg \min_{\tilde{I}} \mathcal{L}([F \circ C] (\tilde{I}), s)$ where $\mathcal{L}$ denotes some types of distance metrics.}

The approaches solving this problem usually exploit the relationship between an input image and its class label for the reconstruction process. Moreover, since the output score $s$ is fixed according to predefined $N$ classes, the reconstruction is limited on images of training subject IDs.

\noindent
\textbf{Definition 2} (Feature Reconstruction). \textit{Given a blackbox functions $F$; and its embedding feature $f = F(I)$ of an unknown image $I$, feature reconstruction is to recover $I$ from $f$ by optimizing $\tilde{I}^* = \arg \min_{\tilde{I}} \mathcal{L}(F (\tilde{I}), f)$}.

\noindent
Compared to the model inversion problem, Feature Reconstruction is more challenging since the constraints on predefined classes are removed. Therefore, the solution for this problem turns into an \textit{\textbf{open-set mode}} where it can reconstruct faces other than the ones used for learning $F$, i.e. face recognition engine.
Moreover, since the parameters of $F$ are inaccessible due to its blackbox setting, directly recovering $I$ based on its gradient is impossible. Therefore, the feature reconstruction task can be reformulated via 
a function (generator) $G: \mathcal{F} \mapsto \mathcal{I}$ as the reverse mapping of $F$.
\begin{equation} \label{eqn:GeneratorFormulation}
\footnotesize
\begin{split}
\tilde{I} &= G(f; \theta_g)\\
\theta_g &= \arg \min_{\theta} \mathbb{E}_{\mathbf{x} \sim p_I} \left[ \mathcal{L}_G^x \left([G \circ F](\mathbf{x}; \theta), \mathbf{x}\right)\right]\\
&= \arg \min_{\theta} \int
\mathcal{L}_G^x \left(\tilde{\mathbf{x}}, \mathbf{x}\right) p_I(\mathbf{x}) d\mathbf{x}\\
\end{split}
\end{equation} 
where $\tilde{\mathbf{x}} = [G \circ F](\mathbf{x}; \theta)$, $\theta_g$ denotes the parameters of $G$, and $p_I(\mathbf{x})$ is the probability density function of $\mathbf{x}$. In other words,  $p_I(\mathbf{x})$ indicates the distribution that image $I$ belonged to (i.e. \textit{the distribution of training data of $F$}).
Intuitively, function $G$ can be seen as a function that maps images from embedding space $\mathcal{F}$ back to image space such that all reconstructed images $[G \circ F](\mathbf{x}; \theta_g)$ are maintained to be close to its real $\mathbf{x}$ with respect to the distance metric $\mathcal{L}_G^x$.  
To produce ``good quality'' synthesis (i.e. \textit{realistic images with ID preservation}), different choices for $\mathcal{L}_G^x$ have been commonly exploited \cite{isola2017image,johnson2016perceptual, mai2018reconstruction} such as pixel difference in image domain via $L_1/ L_2$ distance; Probability Distribution Divergence (i.e. Adversarial loss defined via an additional Discriminator) for realistic penalization; or Perceptual distance that penalizes the image structure in high-level latent space. Among these metrics, except the pixel difference that is computed directly in image domain, the others are indirect metrics where another mapping function (i.e. classifier) from image space to latent space is required.

\subsection{Limitations of Classifier-based Metrics} \label{sec:LimitationClassifierMetric}
Although these indirect metrics have shown their advantages in several tasks, there are limitations when only the blackbox function $F$ and its embedded features are given. 

\noindent
\textit{\textbf{Limitation 1.}} As shown in several adversarial attack works \cite{engstrom2019learning, santurkar2019computer}, 
since the function $F$ is not a one-to-one mapping function from $\mathcal{I}$ to $\mathcal{F}$, it is straightforward to find two images of similar latent representation that are drastically different in image content. Therefore, with no prior knowledge about the subject ID of image $I$, starting to reconstruct it from scratch may easily fall into the case where the reconstructed image $\tilde{I}$ is totally different to $I$ but has similar embedding features.
The current Probability Distribution Divergence with Adversarial Loss or Perceptual Distance is limited in maintaining the constrain ``\textit{the reconstructions of features of the same subject ID should be similar}''.

\noindent
\textit{\textbf{Limitation 2.}} 
Since the access to the structure and intermediate features of $F$ is unavailable in the blackbox mode,
the function $G$ is unable to directly exploit valuable information from the gradient of $F$ and the intermediate representation during embedding. As a result, the distance metrics defined via $F$, i.e. perceptual distance, is less effective as in whitebox setting. 
Next sections will introduce two loss functions to tackle these problems to learn a robust function $G$.

\subsection{Bijective Metrics for Image Reconstruction}
Many metric learning proposals for face recognition \cite{deng2019arcface, LiuNIPS18,liu2017sphereface,  wang2018cosface,wen2016discriminative, zhang2019adacos} have been used to improve both intra-class compactness and inter-class separability with a large margin. 
However, for feature reconstruction, directly adopting these metrics, e.g. angular distance for reconstructed images to cluster images of the same ID is infeasible. 

Therefore, we propose a bijection metric for feature reconstruction task such that the mapping function from image to latent space is one-to-one. The distance between their latent features is equivalent to the distance between images. By this way, these metrics are more aligned to image domain and can be effectively adopted for reconstruction task. Moreover, since two different images cannot be mapped to the same latent features, the metric learning process is more reliable.
The optimization of $G$ in Eqn. \eqref{eqn:GeneratorFormulation} is rewritten as:
\begin{equation} \label{eqn:GeneratorFormulation_Dist}
\footnotesize
\begin{split}
\theta_g &\approx \arg \min_{\theta} \int \mathcal{L}_G^x(\tilde{\mathbf{x}},\mathbf{x}) p_x(\mathbf{x}) d\mathbf{x}
\end{split}
\end{equation}
where $\tilde{\mathbf{x}} = [G \circ F](\mathbf{x}; \theta)$; and $p_x(\mathbf{x})$ denotes a density function estimated from an alternative large-scale face dataset.
Notice that although the access to $p_I(\mathbf{x})$ is not available, this approximation can be practically adopted due to a prior knowledge about $p_I(\mathbf{x})$ that images drawn from $p_I(\mathbf{x})$ are facial images. 
Let $H: \mathcal{I} \mapsto \mathcal{Z}$ define a bijection mapping from $\mathbf{x}$ to a latent variable $\mathbf{z}=H(\mathbf{x})$.
With the bijective property, the optimization in Eqn. \eqref{eqn:GeneratorFormulation_Dist} is equivalent to. 
\begin{equation} \label{eqn:GeneratorFormulation_Bijection}
\footnotesize
\begin{split}
&\arg \min_{\theta} \int \mathcal{L}_G^z(H(\tilde{\mathbf{x}}),H(\mathbf{x})) p_x(\mathbf{x}) d\mathbf{x}\\
=& \arg \min_{\theta} \int \mathcal{L}_G^z(H(\tilde{\mathbf{x}}),H(\mathbf{x})) p_z(\mathbf{z})|\det(\mathbf{J_x^{\top}} \mathbf{J_x})|^{1/2} d\mathbf{z}\\
=& \arg \min_{\theta} \int \mathcal{L}_G^z(\tilde{\mathbf{z}},\mathbf{z}) p_z(\mathbf{z})|\det(\mathbf{J_x^{\top}} \mathbf{J_x})|^{1/2} d\mathbf{z}\\
\end{split}
\end{equation}
where $\title{\mathbf{z}}=H(\tilde{\mathbf{x}})$; $p_x(\mathbf{x})=p_z(\mathbf{z})|\det(\mathbf{J_x^{\top}} \mathbf{J_x})|^{1/2}$ by the change of variable formula; $\mathbf{J_x}$ is the Jacobian of $H$ with respect to $\mathbf{x}$; and $\mathcal{L}_G^z$ is the distance metric in $\mathcal{Z}$. Intuitively, Eqn. \eqref{eqn:GeneratorFormulation_Bijection} indicates that instead of computing the distance $\mathcal{L}_G^x$ and estimating $p_x(\mathbf{x})$ directly in image domain, the optimization process can be equivalently accomplished via the distance $\mathcal{L}_G^z$ and density $p_z(\mathbf{z})$ in $\mathcal{Z}$ according to the bijective property of $H$.

\noindent
\textbf{\textit{The prior distributions $p_z$.}} In general, there are various choices for the prior distribution $p_z$ and the ideal one should have two properties: (1) \textit{simplicity in density estimation}, and (2) \textit{easily sampling}. Motivated from these properties, we choose Gaussian distribution for $p_z$. Notice that other distribution types are still applicable in our framework.

\noindent
\textbf{\textit{The distance metric $\mathcal{L}_G^z$.}}
With the choice of $p_z$ as a Gaussian, 
the distance between images in $\mathcal{I}$ is equivalent to the deviation between Gaussians in latent space. Therefore, we can effectively define $\mathcal{L}_G^z$  as the squared Wasserstein coupling distance between two Gaussian distributions.
\begin{equation}
\small
\begin{split}
\mathcal{L}_G^z(\tilde{\mathbf{z}},\mathbf{z})
=& d(\tilde{\mathbf{z}},\mathbf{z})
= \inf \mathbb{E}(||\mathbf{\tilde{z}}-\mathbf{z}||_2^2)\\
=& || \tilde{\mu} - \mu ||^2_2 
+ \text{Tr}(\tilde{\Sigma} + \Sigma - 2(\tilde{\Sigma}^{1/2}\Sigma\tilde{\Sigma}^{1/2})^{1/2})
\end{split}
\end{equation}
where $\{\tilde{\mu}, \tilde{\Sigma}\}$ and $\{\mu, \Sigma\}$ are the means and covariances of $\mathbf{\tilde{z}}$ and $\mathbf{z}$, respectively.
The metric $\mathcal{L}_G^z$ then can be extended with image labels to  reduce the distance between images of the same ID and enhance the margin between different IDs.
\begin{equation}
\small
\mathcal{L}_G^{z_{id}}(\mathbf{\tilde{z}_1},\mathbf{\tilde{z}_2}) = 
\begin{cases}
d(\mathbf{\tilde{z}_1},\mathbf{\tilde{z}_2})       &  \text{if } l_{\mathbf{\tilde{z}_1}} = l_{\mathbf{\tilde{z}_2}}\\
\max(0,m-d(\mathbf{\tilde{z}_1},\mathbf{\tilde{z}_2}) )  & \text{if } l_{\mathbf{\tilde{z}_1}} \neq l_{\mathbf{\tilde{z}_2}}
\end{cases}
\end{equation}
where $m$ defines parameter controlling the margin between classes; and  $\{l_{\mathbf{\tilde{z}_1}}, l_{\mathbf{\tilde{z}_2}}\}$ denote the subject ID of $\{\mathbf{\tilde{z}_1}, \mathbf{\tilde{z}_2}\}$.

\begin{figure}[t]
	\centering \includegraphics[width=0.9\columnwidth]{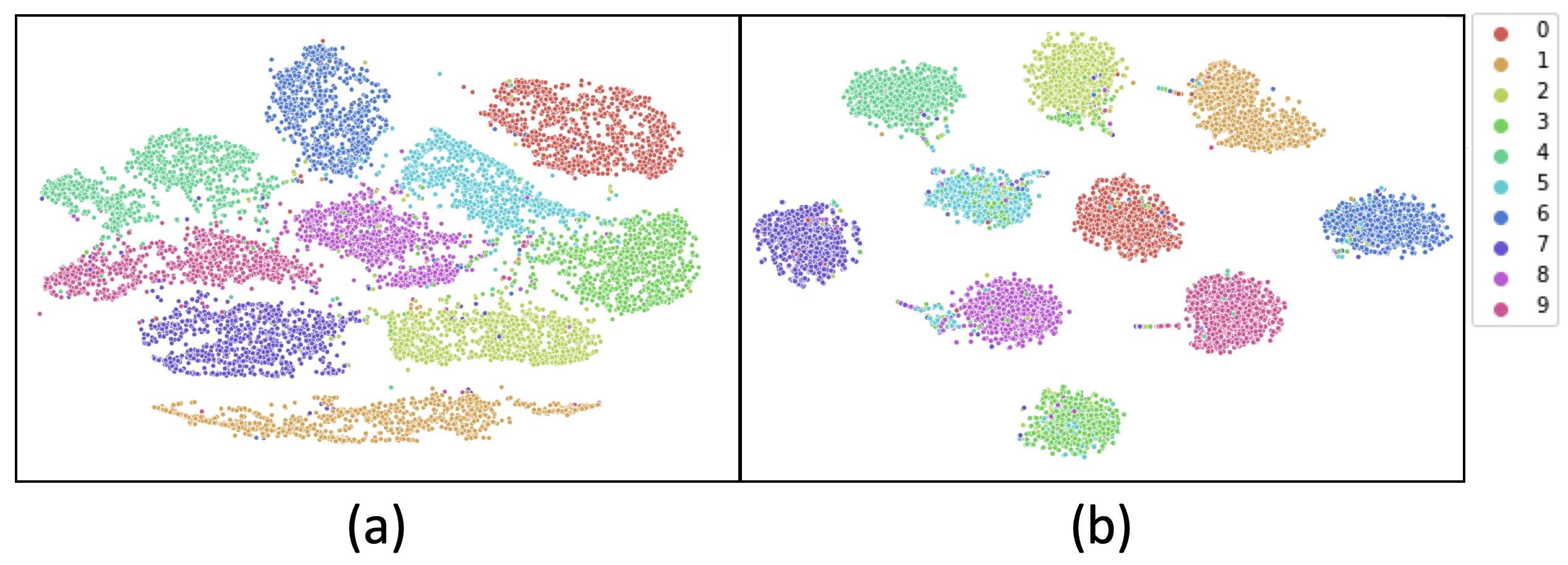}
	\caption{\textbf{The distributions of synthesized MNIST samples} on testing set (a) without, and (b) with adopting Bijective Metric.}
	\label{fig:MNISTDistribution}
	\vspace{-5mm}
\end{figure}

\noindent
\textbf{\textit{Learning the Bijection $H$.}} In order to effectively learn the bijection $H$, we adopt the structure of mapping function from \cite{dinh2016density, Duong_2017_ICCV,duong2019learning} as the backbone for the tractable log-det computation with the log-likelihood loss for training process. Moreover, to further improve the discriminative property of $H$ in latent space $\mathcal{Z}$, we propose to exploit the ID label in training process of $H$. Particularly, given $K$ classes (i.e. ID) of the training set, we choose $K$ Gaussian distributions with different means $\{\mu_1, \mu_2, .., \mu_K\}$ and covariances $\{\Sigma_1,\Sigma_2,...,\Sigma_K\}$ and enforce samples of each class distributed on its own prior distribution, i.e. $\mathbf{z}_k \sim \mathcal{N}(\mu_k, \Sigma_k)$.
Formally, the log-likelihood loss function to learn $H$ is formulated as $\theta_H^* = \arg \max_{\theta_H} \log p_x(\mathbf{x}, k;\theta_H)= \arg \max_{\theta_H} \log p_z(\mathbf{z}, k;\theta_H) + \frac{1}{2}\log |\det(\mathbf{J_x^{\top}} \mathbf{J_x})|$. 

\subsection{Reconstruction from Distillation Knowledge}
In the simplest approach, the generator $G$ can still learn to reconstruct image by adopting the Perceptual Distance as in previous works to compare $F(\tilde{I})$ and $F(I)$. However, as mentioned in Sec. \ref{sec:LimitationClassifierMetric}, due to limited information that can be accessed from $F$, ``key'' information (i.e. the gradients of $F$ as well as its intermediate representations) making the perceptual loss effective is lacking.
Therefore, we propose to first distill the knowledge from the blackbox $F$ to a ``student'' function $F^S$ and then take advantages of these knowledge via $F^S$ for training the generator. On one hand, via the distillation process, $F^S$ can mimic $F$ by aligning its feature space to that of $F$ and keeping the semantics of the extracted features for reconstruction. On the other hand, with $F^S$, the knowledge about the embedding process of $F$ (i.e. gradient, and intermediate representation) becomes more transparent; and, therefore, maximize the information which can be exploited from $F$.
Particularly, let $F^S: \mathcal{I} \mapsto \mathcal{F}$ and $F^S=F^S_1 \circ F^S_2 \cdots \circ F^S_n$ be the composition of $n$-sub components. The knowledge from $F$ can be distilled to $F^S$ by aligning their extracted features as.
\begin{equation} \label{eqn:LearningStudentMatcher}
\footnotesize
\begin{split}
\theta_S =& \arg \min_{\theta_S} \mathcal{L}_S = \mathbb{E}_{\mathbf{x} \sim p_x} d_{distill}\left(F(\mathbf{x}),F_S(\mathbf{x}; \theta_S) \right)\\
=& \arg \min_{\theta_S} \mathbb{E}_{\mathbf{x} \sim p_x} \left\| 1 - \frac{F(\mathbf{x})}{\parallel F(\mathbf{x})\parallel}* \frac{F^S(\mathbf{x}; \theta_S)}{\parallel F^S(\mathbf{x}; \theta_S)\parallel}\right\|^2_2
\end{split}
\end{equation}
Then $G$ is enhanced via the distilled knowledge of both final embedding features and intermediate representation by.
\begin{equation} \label{eqn:DistillationLoss}
\footnotesize
\begin{split}
\mathcal{L}_G^{distill}(\mathbf{\tilde{x}}, \mathbf{x}) &= \sum_{j=1}^n \lambda_j \frac{\left\| F_j^S(\mathbf{\tilde{x}};\theta_S) - F_j^S(\mathbf{x};\theta_S) \right\|}{W_j H_j C_j} \\
+& \lambda_a \left\| 1 - \frac{F^S(\mathbf{\tilde{x}};\theta_S)}{\parallel F^S(\mathbf{\tilde{x}};\theta_S)\parallel}* \frac{F^S(\mathbf{x}; \theta_S)}{\parallel F^S(\mathbf{x}; \theta_S)\parallel}\right\|^2_2
\end{split}
\end{equation}
where $\{\lambda_j\}^n_1$ and $\lambda_a$ denote the hyper-parameters controlling the balance between terms. The first component of $\mathcal{L}_G^{distill}(\mathbf{\tilde{x}}, \mathbf{x})$ aims to penalize the differences between the intermediate structure of the desired and reconstructed facial images while the second component validates the similarity of their final features.

\begin{figure*}[t]
	\centering \includegraphics[width=1.8\columnwidth]{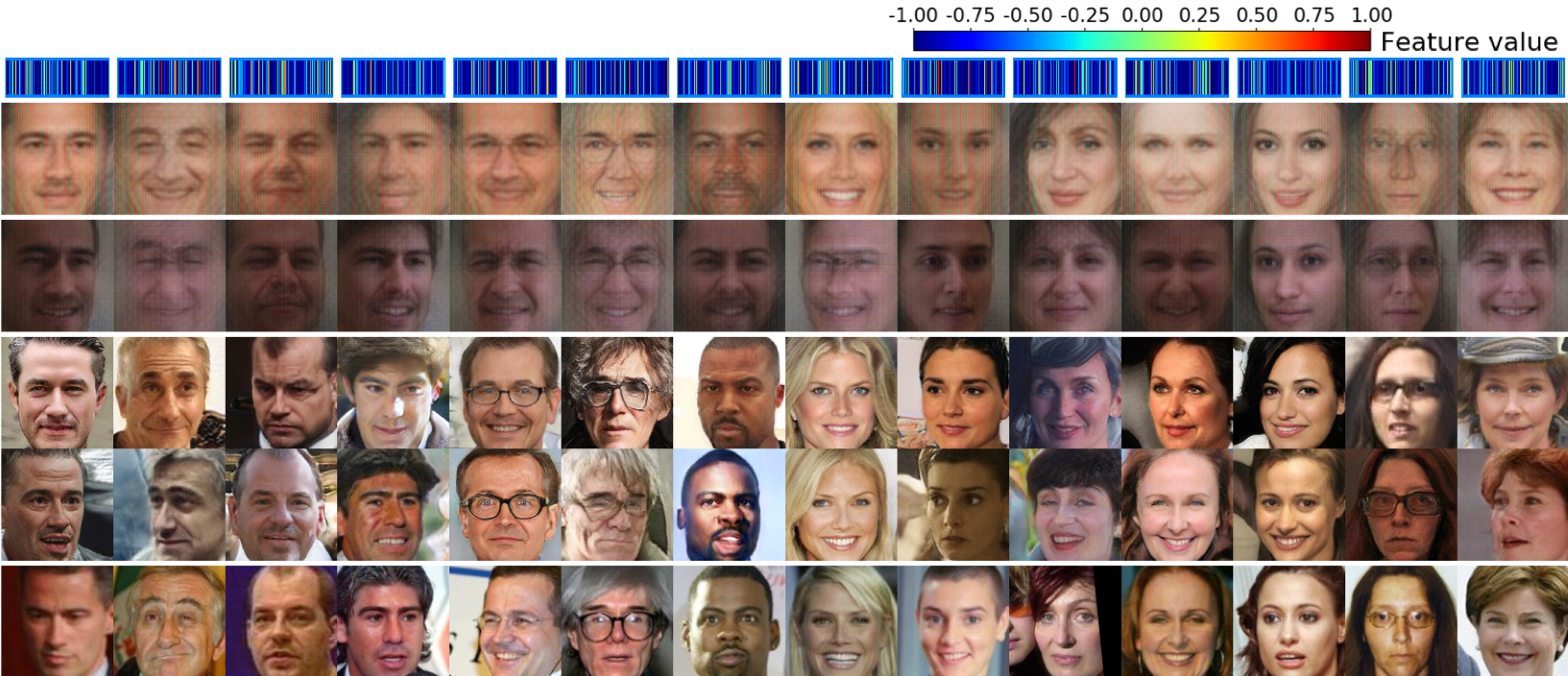}
	\caption{\textbf{Feature Reconstruction against in-the-wild facial variations.} For each subject, given an input feature (1st row), while  VGG-NBNet \cite{mai2018reconstruction} and MPIE-NBNet \cite{mai2018reconstruction} (2nd and 3rd rows) reconstruct faces with limited quality, DibiGAN in whitebox (4th row) and blackbox (5th row) modes are able to produce realistic faces with better ID preservation comparable to real faces (6th row).} 
	\label{fig:LFWRecon}
	\vspace{-5mm}
\end{figure*}

\subsection{Learning the Generator} \label{sec:LearningStategies}
Fig. \ref{fig:ProposedFramework} illustrates our proposed framework with Bijective Metric and Distillation Process to learn the generator $G$.

\noindent
\textbf{Network Architecture.} 
Given an input image $\mathbf{x}$, the generator $G$ takes $F(\mathbf{x})$ as its input and aims to synthesize an image $\tilde{\mathbf{x}}$ that is as similar to $\mathbf{x}$ as possible in terms of identity and appearance. 
We adopt the GAN-based generator structure for $G$ and optimize using different criteria.
\begin{equation}
\footnotesize
\begin{split}
\mathcal{L}_G =& \lambda_{b} \mathcal{L}^{biject} + \lambda_d \mathcal{L}^{distill} + \lambda_{adv} \mathcal{L}^{adv} + \lambda_{r} \mathcal{L}^{recon}\\
\mathcal{L}^{biject} =& \mathbb{E}_{\mathbf{x}\sim p_x} \left[ \mathcal{L}_G^x \left([G \circ F](\mathbf{x}; \theta), \mathbf{x}\right)\right]\\
+& \mathbb{E}_{\mathbf{x}_1,\mathbf{x}_2\sim p_x} \left[ \mathcal{L}_G^{x_{id}} \left([G \circ F](\mathbf{x}_1; \theta), [G \circ F](\mathbf{x}_2;\theta\right)\right]\\
=&\mathbb{E}_{\mathbf{z}\sim p_z} \left[ \mathcal{L}_G^z \left(\mathbf{\tilde{z}}, \mathbf{z}\right)\right] + \mathbb{E}_{\mathbf{z}_1,\mathbf{z}_2\sim p_z} \left[ \mathcal{L}_G^{z_{id}} \left(\mathbf{\tilde{z}}_1, \mathbf{\tilde{z}}_2\right)\right]\\
\mathcal{L}^{distill} =& \mathbb{E}_{\mathbf{x}\sim p_x} \left[ \mathcal{L}_G^{distill} \left([G \circ F](\mathbf{x}; \theta), \mathbf{x}\right)\right]\\
\mathcal{L}^{adv} =& \mathbb{E}_{\mathbf{x}\sim p_x} \left[ D\big([G \circ F](\mathbf{x};\theta)\big)\right]\\
\mathcal{L}^{recon} =& \mathbb{E}_{\mathbf{x}\sim p_x} \left[ \big\| [G \circ F](\mathbf{x}) - \mathbf{x} \big\|_1\right]
\end{split}
\end{equation}
where $\{\mathcal{L}^{biject},\mathcal{L}^{distill}, \mathcal{L}^{adv}, \mathcal{L}^{recon}\}$ denote the bijective, distillation, adversarial, and reconstruction losses, respectively. $\{\lambda_{b}, \lambda_d, \lambda_{adv}, \lambda_{r}\}$ are their parameters controlling their relative importance. $D$ is a discriminator distinguishing the real images from a synthesized one. 
There are three main critical components in our framework including the Bijective $H$, the student matcher $F^S$ for ID preservation; and the discriminator $D$ for realistic penalization. 
The Discriminator $D$ is updated with the objective function as.
\begin{equation}
\footnotesize
\begin{split}
\mathcal{L}_D =& \mathbb{E}_{\mathbf{x} \sim p_{x}} \left[ D\big([G \circ F](\mathbf{x};\theta)\big) \right] - \mathbb{E}_{\mathbf{x} \sim p_{x}} \left[ D\big(\mathbf{x})\big) \right]\\
+& \lambda \mathbb{E}_{\mathbf{\hat{x}} \sim p_{\hat{x}}} \left[ \left( \|\nabla_{\mathbf{\hat{x}}}D(\mathbf{\hat{x}})\|_2 -1 \right)^2\right]
\end{split}
\end{equation}
where $p_{\hat{x}}$ is the random interpolation distribution 
between real and generated images \cite{gulrajani2017improved}.
Then, the whole framework is trained following GAN-based minimax strategy.

\noindent
\textbf{Learning Strategies.} 
Besides the losses, we introduce a Feature-Conditional Structure for $G$ and a exponential Weighting Strategy to adaptively scheduling the importance factors between loss terms during training process.

\noindent
\textbf{\textit{Feature-conditional Structure.}} A natural design for the structure of $G$ is 
to directly use $F(\mathbf{x})$ as the input for $G$. However, this structure limits the learning capability of $G$. Particularly, besides ID information, $F(\mathbf{x})$ may include other ``background'' conditions such as poses, illuminations, expressions.
Therefore, setting $F(\mathbf{x})$ as the only input implicitly enforces $G$ to ``strictly'' model these factors as well.
This makes the training process of $G$ less effective.
To relax this constraint, we introduce a Feature-Conditional structure 
(i.e. the generator structure in Fig. \ref{fig:ProposedFramework}) 
where a random variable $\mathbf{v}$ is adopted as an additional input so that these background factors can be modeled through $\mathbf{v}$. Moreover, we propose to use $\mathbf{v}$ as the direct input to $G$ and inject the information from $F(\mathbf{x})$ through out the structure. By this way, $F(\mathbf{x})$ can act as the conditional ID-related information for all reconstruction scales and gives the better synthesis.

\noindent
\textbf{\textit{Exponential Weighting Strategy.}}
As the progressive growing training strategy \cite{karras2017progressive} initializes its learning process on synthesizing low-resolution images and then gradually increasing their levels of details, it is quite effective for enhancing the details of generated images in general.
However, this strategy has limited capability in preserving the subject ID. In particular, in the early stages at low scales with blurry synthesis, it is difficult to control the subject ID of faces to be synthesized while in the later stages at higher scales when the generator 
becomes more mature and 
learns to add more details, the IDs of those faces have already been constructed and become hard to be changed. Therefore, we propose to adopt a exponential weighting scheme for (1) emphasizing on ID preservation in  early stages; and (2) enhancing the realism in later stages. 
Particularly, the parameter set $\{\lambda_{b}, \lambda_d, \lambda_{adv}, \lambda_{r}\}$ is set to $\lambda_{b}=\alpha e^{R_M - R(i)},\lambda_{d}=e^{R_M - R(i)}, \lambda_{adv}=\beta e^{R(i)}, \lambda_{r}=e^{R_M - R(i)}$ where $R(i)$ denotes the current scales of stage $i$ and $R_M$ is the maximum scales to be learned by $G$.

\vspace{-2mm}
\section{Experimental Results}
\vspace{-1mm}
We qualitatively and quantitatively validate our proposed method in both reconstructed quality and ID preservation in several in-the-wild modes such as poses, expressions, and occlusions. Both image-quality
and face-matching datasets
are used for evaluations. 
Different face recognition engines are adopted to demonstrate the robustness of our model.

\noindent
\textbf{Data Setting.}
Our training data includes the publicly available Casia-WebFace \cite{yi2014learning} with 490K labeled facial images of over 10K subjects. The duplicated subjects between training and testing sets are removed to ensure no overlapping between them.
For validation, as commonly used for attribute learning and image quality evaluation, we adopt the testing split of 10K images from CelebA \cite{liu2015faceattributes} to validate the reconstruction quality. For ID preservation, we explore LFW \cite{huang2008labeled}, AgeDB \cite{moschoglou2017agedb}, and CFP-FP \cite{sengupta2016frontal} which provide face verification protocols against different in-the-wild face variations.
Since each face matcher engine requires different preprocessing process, the training and testing data are aligned to the required template accordingly. 

\noindent
\textbf{Network Architectures.} We exploited the Generator structure of PO-GAN \cite{karras2017progressive} with 5 convolutional blocks for $G$ while the Feature Conditional branch consists of 8 fully connected layers. The discriminator $D$ includes five consecutive blocks of two convolution and one downsampling operators. In the last block of $D$, the minibatch-stddev operator followed by convolution and fully connected are also adopted. AdaIN operator \cite{huang2017arbitrary} is applied for feature injection node.
For the bijection $H$, we set a configuration of 5 sub-mapping functions 
where each of them is presented with two 32-feature-map residual blocks. This structure is trained using the log-likelihood objective function on Casia-WebFace. 
Resnet-50 \cite{he2016deep} is adopted for $F^S$.

\noindent
\textbf{Model Configurations.} Our framework is implemented in TensorFlow and all the models are trained on a machine with four NVIDIA P6000 GPUs. 
The batch size is set based on the resolution of output images, for the very first resolution of output images ($4 \times 4$), the batch size is set to $128$, the batch size will be divided by two when the resolution of images is doubled.
We use Adam Optimizer 
with the started learning rate of $0.0015$. 
We experimentally set $\{\alpha = 0.001, \beta = 1.0, \lambda_j=1, \lambda_a = 10.0\}$
to maintain the balanced values between loss terms.

\begin{figure*}[t]
	\centering \includegraphics[width=1.82\columnwidth]{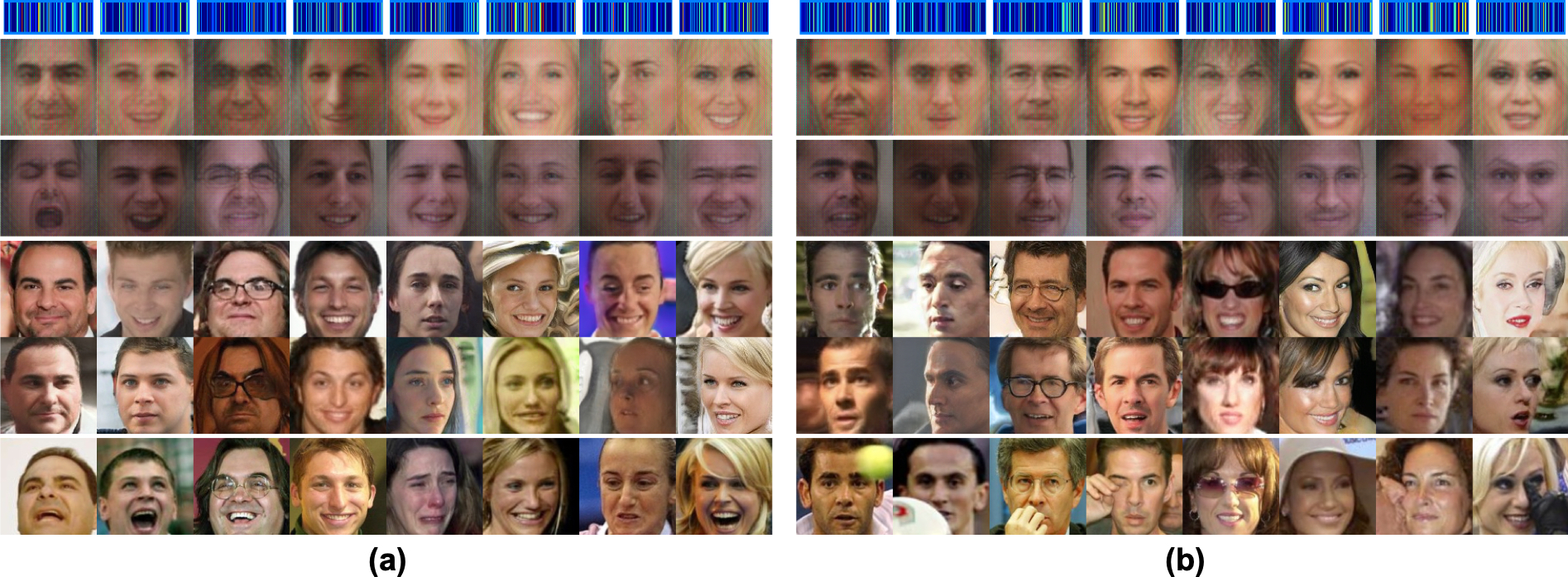}
	\caption{\textbf{Feature Reconstruction against expressions (a) and occlusions (b).} For each subject, the 1st row shows the input feature. The next five rows are VGG-NBNet \cite{mai2018reconstruction}, MPIE-NBNet \cite{mai2018reconstruction}, Our DibiGAN in whitebox and blackbox settings, and Real Faces, respectively.}
	\label{fig:ReconExpressionOcclusion}
	\vspace{-5mm}
\end{figure*}

\begin{figure}[t]
	\centering \includegraphics[width=0.84\columnwidth]{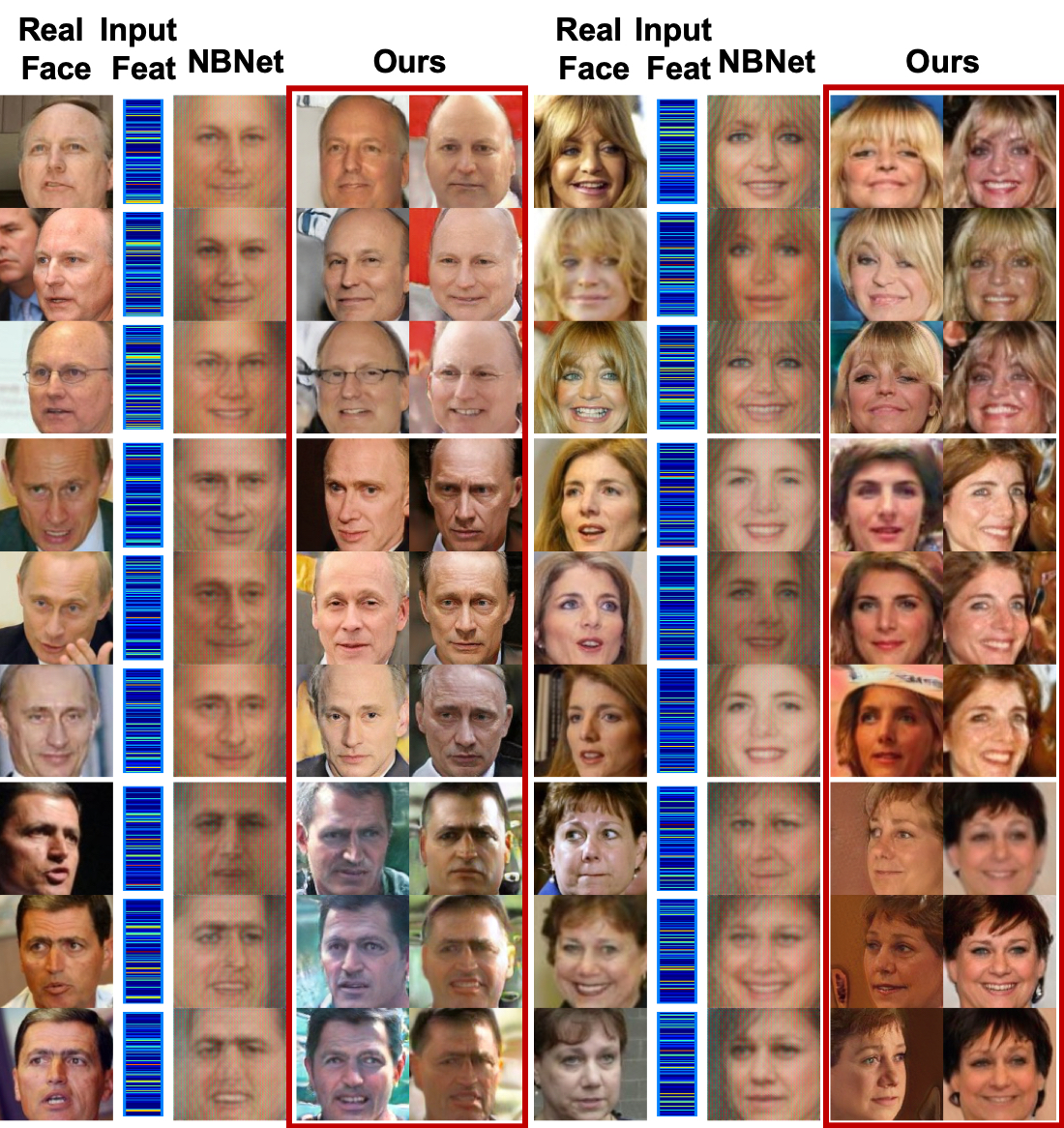}
	\caption{\textbf{Feature Reconstruction against features of the same subject.} For each subject, the first and second columns show different real faces and their features of a subject. Compared to VGG-NBNet \cite{mai2018reconstruction} (third column), our DibiGAN in whitebox and blackbox modes can effectively preserve the ID of the subjects.}
	\label{fig:ReconSameSubject}
	\vspace{-6mm}
\end{figure}

\noindent
\textbf{Ablation Study.}
To study the effectiveness of the proposed bijective metric for image reconstruction task, we employ an ablation study on MNIST \cite{lecun1998mnist} 
with LeNet \cite{lecun1998gradient} as the function $F$. 
We also set to whitebox mode where $F$ is directly used in $\mathcal{L}^{distill}$ to remove the effects of other factors. Then 50K training images from MNIST and their $1 \times 1024$ feature vectors  are used to train $G$. Notice that since the image size is $32 \times 32$, $G$ and $D$ structures are configured with three convolutional blocks. The resulting distributions of synthesized testing images of all classes without and with $\mathcal{L}^{biject}$ are plotted in Fig. \ref{fig:MNISTDistribution}. Compared to $G$ learned with only classifier-based metrics (Fig. \ref{fig:MNISTDistribution}(a)), the one with bijective metric learning (Fig. \ref{fig:MNISTDistribution}(b)) is supervised with more direct metric learning mechanism in image domain, and, 
therefore, shows the advantages with enhanced intra- and inter-class distributions.

\vspace{-1mm}
\subsection{Face Reconstruction Results}
\vspace{-1mm}
This section demonstrates the capability of our proposed methods in terms of effectively synthesizing faces from subject's features.
To train DibiGAN, we adopt the ArcFace-Resnet100 \cite{deng2019arcface} trained on 5.8M images of 85K subjects for function $F$ and extract the $1 \times 512$ feature vectors for all training images. These features together with the training images are then used to train the whole framework. We divided the experiments in two settings, i.e. \textit{whitebox} and \textit{blackbox}, where the main difference is the visibility of the matcher structure during training process. In the whitebox mode $F$ is directly used in Eqn. \eqref{eqn:DistillationLoss} to evaluate $\mathcal{L}_G^{distill}$ while in the blackbox mode, $F^S$ is learned from $F$ through a distillation process as in Eqn. \eqref{eqn:LearningStudentMatcher} and used for $\mathcal{L}_G^{distill}$.
The first row of Table \ref{tab:MatchingAccuracyAll} shows the matching accuracy of $F$ and $F^S$ using real faces on benchmarking datasets. 

\noindent
\textbf{Face Reconstruction from features of frontal faces.}
After training, given only the deep features extracted from $F$ on testing images, the generator $G$ is applied to synthesize the subjects' faces.
Qualitative examples of our synthesized faces
in comparison with other methods are illustrated in Fig. \ref{fig:LFWRecon}. 
As can be seen, our generator $G$ is able to reconstruct realistic faces even when their embedding features are extracted from faces with a wide range of in-the-wild variations. More importantly, in both whitebox and blackbox settings, our proposed method successfully preserves the ID features of these subjects. In whitebox setting, since the structure of $F$ is accessible, the learning process can effectively exploit different aspects of embedding process from $F$ and produce a generator $G$ that depicts better facial features of the real faces. For example, together with ID information, poses, glasses, or hair style from the real faces can also be recovered. On the other hand, although the accessible information is very limited in blackbox setting, the learned $G$ can still be enjoyed from the distilled knowledge of $F^S$ and effectively fill the knowledge gap with whitebox setting. In comparison to different configurations of NBNet \cite{mai2018reconstruction}, better faces in terms of image quality and ID preservation can be obtained by our proposed model.

\begin{figure}[t]
	\centering \includegraphics[width=0.9\columnwidth]{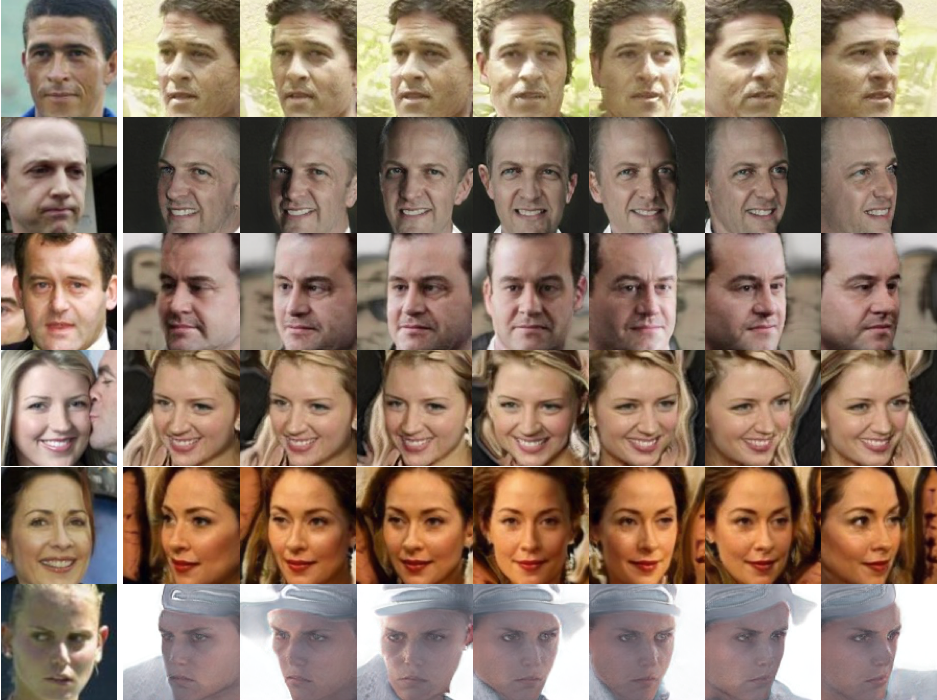}
	\caption{From the input features, our model can synthesize various conditions of a face by varying the ``background'' variable $\mathbf{v}$.}
	\label{fig:ReconDifferentPoses}
	\vspace{-6mm}
\end{figure}

\begin{table*}[!t] 
	\small
	\centering
	\caption{\textbf{Realism Quality and Matching Accuracy.} Comparison results in Multi-Scale Structural Similarity (MS-SSIM) (\textit{the smaller value is better}); Inception score and Matching Accuracy (\textit{the higher value is better}). For each configuration in (A)-(C) and (D)-(F), each loss function is cumulative enable on the top of the previous configuration. $-$ denotes ``not applicable''.
	}
	\label{tab:MatchingAccuracyAll}
	\begin{tabular}{|l|cc|ccc|cc|ccc|}
		\hline
		\multirow{3}{*}{}&\multicolumn{5}{c|}{\textbf{White-box Reconstruction}} & \multicolumn{5}{c|}{\textbf{Black-box Reconstruction}}\\
		\cline{2-11}
		\multirow{3}{*}{} & \multicolumn{2}{c|}{\textbf{CelebA}} & \multirow{2}{*}{\textbf{LFW}} & \multirow{2}{*}{\textbf{AgeDB}} & \multirow{2}{*}{\textbf{CFP-FP}} & \multicolumn{2}{c|}{\textbf{CelebA}}& \multirow{2}{*}{\textbf{LFW}} & \multirow{2}{*}{\textbf{AgeDB}} & \multirow{2}{*}{\textbf{CFP-FP}} \\
		\multirow{3}{*}{} & MS-SSIM & IS & \multirow{2}{*}{} & \multirow{2}{*}{} & \multirow{2}{*}{} & MS-SSIM & IS & \multirow{2}{*}{} & \multirow{2}{*}{} & \multirow{2}{*}{} \\
		\hline
		Real Faces \footnotemark       & 0.305 & 3.008    & 99.78\%    & 98.40\%     & 97.10\% & 0.305 & 3.008 & 99.70\%  & 96.80\%     & 93.10\%\\ 
		\hline
		VGG-NBNet \cite{mai2018reconstruction}        & $-$ & $-$ & $-$    & $-$     & $-$ & 0.661 & 1.387  & 91.42\%   & 80.42\%     & 74.63\%\\
		MPIE-NBNet \cite{mai2018reconstruction}        & $-$ & $-$ & $-$    & $-$     & $-$ & 0.592 & 1.484    & 93.17\%    & 79.45\%     & 78.51\%\\
		\hline
		(A) PO\_GAN \cite{karras2017progressive}    &\textbf{0.331}& \textbf{2.226}    & 68.20\%    & 63.42\%     & 68.89\%  & \textbf{0.315} & 2.227    & 66.63\%    & 62.37\%     & 65.59\%\\
		(B) \quad + $\mathcal{L}^{distill}$     & 0.343& 2.073   &  96.03\%  & 83.33\%    & 79.07\% &  0.337 & \textbf{2.238}& 94.95\%    & 81.56\%  &  78.80\%\\
		(C) \quad + $\mathcal{L}^{biject}$     & 0.358& 2.052   & \textbf{98.1\%}    & \textbf{88.16\%}     & \textbf{88.01\%}  & 0.360&2.176    & \textbf{97.30\%}    & \textbf{85.71\%}     & \textbf{82.51\%}\\
		\hline
		\hline		
		(D) \textbf{Ours}     & 0.316 & 2.343  & 79.82\%    & 77.20\%     & 81.71\%  & 0.305 & \textbf{2.463}   & 77.57\%   & 76.83\%     & 80.66\%\\
		(E) \quad + $\mathcal{L}^{distill}$     & \textbf{0.306}& 2.349    & 97.76\%    & 92.33\%     & 89.20\%  & 0.305 & 2.423 & 97.06\%    & 91.70\%     & 84.83\%\\
		(F) \quad + $\mathcal{L}^{biject}$     & 0.310& \textbf{2.531}    & \textbf{99.18\%}    & \textbf{94.18\%}     & \textbf{92.67\%}  &\textbf{ 0.303} & 2.422    & \textbf{99.13\%}    & \textbf{93.53\%}     & \textbf{89.03\%}\\
		\hline
	\end{tabular}
	\vspace{-5mm}
\end{table*}

\noindent
\textbf{Effect of expressions and occluded regions.} Fig. \ref{fig:ReconExpressionOcclusion} illustrates our synthesis from features of faces that contain both expressions and occlusions. Similar to previous experiment, our model robustly depicts realistic faces with similar ID features as in the real faces. 
Those reconstructed faces' quality again outperforms NBNet in both realistic and ID terms. 
Notice that the success of robustly handling with those challenging factors comes from two properties: (1) The matcher $F$ was trained to ignore those facial variations in its embedding features; and (2) both bijective metric learning and distillation process can effectively exploit necessary knowledge from $F$ as well as real face distributions in image domain for synthesis process.

\noindent
\textbf{Effect of different features from the same subject.} Fig. \ref{fig:ReconSameSubject} illustrates the advantages of our method in synthesizing faces given different feature representations of the same subject. These results further show the advantages of the proposed bijective metric in enhancing the boundary between classes and constrain the similarity between reconstructed faces of the same subject in image domain. As a result, reconstructed faces from features of the same subject ID not only keep the features of that subject (i.e. similar to real faces) but also share similar features among each other.

\noindent
\textbf{Effect of random variable $\mathbf{v}$.}
As mentioned in Sec. \ref{sec:LearningStategies}, the variable $\mathbf{v}$ is incorporated to model background factors so that $G$ can be more focused on modeling ID features.
Therefore, by fixing the input feature and varying this variable values, different conditions of that face can be synthesized as shown in Fig. \ref{fig:ReconDifferentPoses}. These results further illustrate the advantages of our model structure in its capability of capturing various factors for the reconstruction process.

\vspace{-1mm}
\subsection{Face Quality and Verification Accuracy}
\vspace{-1mm}
In order to quantitatively validate the realism of our reconstructed images and how well they can preserve the ID of the subjects, three metrics are adopted: (1) Multi-scale Structural similarity (MS-SSIM) \cite{odena2017conditional}; (2) Inception Score (IS) \cite{salimans2016improved}; and (3) face verification accuracy.

\noindent
\textbf{Image quality.} To quantify the realism of the reconstructed faces, we synthesize testing images of CelebA in several training configurations as shown in Table \ref{tab:MatchingAccuracyAll}, where each loss function in cumulatively enables on the top of the previous configuration. Then MS-SSIM and IS metrics are applied to measure their image quality. 
We also compare our model in both whitebox and blackbox settings with other baselines including PO\_GAN structure \cite{karras2017progressive} and NBNet \cite{mai2018reconstruction}. Notice that we only adopt the adversarial and reconstruction losses for configs (A) and (D). For all configs (A), (B), and (C), PO\_GAN baseline takes only the embedding features as its input.
These results show that 
in all configurations, our method maintains comparative reconstruction quality as PO\_GAN and very close to that of real faces. Moreover, our synthesis consistently outperforms NBNet in both metrics.

\footnotetext{We report the accuracy of original matcher $F$ for whitebox setting and $F^S$ for blackbox setting.}

\noindent
\textbf{ID Preservation.} Our model is experimented against 
LFW, AgeDB, and CFP-FP where an image in each positive pair is substituted by the reconstructed one while the remaining image of that pair is kept as the reference real face. 
The matching accuracy 
is reported in Table \ref{tab:MatchingAccuracyAll}. These results further demonstrate the advantages and contributions of each component in our framework. 
Compared to PO\_GAN structure, our Feature-Conditional Structure gives more flexibility in modeling ID features, and achieves better matching accuracy. In combination with distilled knowledge from $F^S$, the Generator produces a big jump in accuracy and close the gap to real faces to only 2.02\% and 2.72\% on LFW in whitebox and blackbox settings, respectively. By further incorporating the bijective metric, these gaps are further reduced to only 0.6\% and 0.65\% for the two settings.

\begin{table} [!t]
	\small
	\centering
	\caption{Accuracy against different blackbox face matchers.} 
	\begin{tabular}{|l|c|c|c|}
		\hline
		\textbf{Matcher} & \textbf{LFW} & \textbf{AgeDB} & \textbf{CFP-FP}\\
		\hline
		\hline
		ArcFace\cite{deng2019arcface}-Real & 99.78\% & 98.40\% & 97.1\%\\
		\textbf{ArcFace-Recon} & \textbf{99.13\%} & \textbf{93.53\%} & \textbf{89.03\%}\\
		\hline
		FaceNet\cite{schroff2015facenet}-Real & 99.55\% &  90.16\%& 94.05\%\\
		\textbf{FaceNet-Recon} & \textbf{98.05\%} & \textbf{89.80\%} & \textbf{87.19\%}\\
		\hline
		SphereFacePlus\cite{LiuNIPS18}-Real & 98.92\% & 91.92\% & 91.16\%\\
		\textbf{SphereFacePlus-Recon} & \textbf{97.21\%} & \textbf{88.98\%} & \textbf{86.86\%} \\
		\hline
	\end{tabular}\label{table:BlackboxDifferentMatcher}
	\vspace{-6mm}
\end{table}

\noindent
\textbf{Reconstructions against different Face Recognition Engines.} To illustrate the accuracy of our propose structure, we validate the its performance against different 
face recognition engines as shown in Table \ref{table:BlackboxDifferentMatcher}. All Generators are set to blackbox mode and only the final extracted features are accessible. Our reconstructed faces are able to maintain the ID information and achieve competitive accuracy as real faces. 
These performance again emphasizes the accuracy of our model in capturing behaviours of the feature extraction functions $F$ and provides high quality reconstructions.

\vspace{-2mm}
\section{Conclusions.}
\vspace{-1mm}
This work has presented a novel generative structure with Bijective Metric Learning for feature reconstruction problem to unveil the subjects' faces given their deep blackboxed features. Thanks to the introduced Bijective Metric and Distillation Knowledge, our DibiGAN effectively maximizes the information to be exploited from a given blackbox face matcher. Experiments on a wide range of in-the-wild face variations against different face matching engines demonstrated the advantages of our method on synthesizing realistic faces with subject's visual identity.

{\small
\bibliographystyle{ieee_fullname}
\bibliography{egbib}
}

\end{document}